\documentclass[12pt]{elsarticle}

\usepackage{lipsum}
\makeatletter
\def\ps@pprintTitle{%
 \let\@oddhead\@empty
 \let\@evenhead\@empty
 \def\@oddfoot{}%
 \let\@evenfoot\@oddfoot}
\makeatother

\usepackage{graphicx}
\usepackage{bbm}
\usepackage{hyperref}
\usepackage{amssymb}
\usepackage{siunitx}
\newcommand{\Ie}{\textit{I}.\textit{e}.}
\newcommand{\ie}{\textit{i}.\textit{e}.}

\newcommand{\Eg}{\textit{E}.\textit{g}.}

\usepackage{lineno}

\begin{document}

\begin{frontmatter}


\title{DeadNet: Identifying Phototoxicity from Label-free Microscopy Images of Cells using Deep ConvNets}

\author[IDAC]{David Richmond\fnref{label2}}
\author[NIC]{Anna Payne-Tobin Jost\fnref{label2}}
\author[NIC]{Talley Lambert}
\author[NIC]{Jennifer Waters}
\author[IDAC]{Hunter Elliott}


\address[IDAC]{Image and Data Analysis Core (IDAC), and}
\address[NIC]{Nikon Imaging Center (NIC), \\ Harvard Medical School, Boston, MA}

\fntext[label2]{These authors contributed equally to this work}



\begin{abstract}
Exposure to intense illumination light is an unavoidable consequence of fluorescence microscopy, and poses a risk to the health of the sample in every live-cell fluorescence microscopy experiment.
Furthermore, the possible side-effects of phototoxicity on the scientific conclusions that are drawn from an imaging experiment are often unaccounted for.
Previously, controlling for phototoxicity in imaging experiments required additional labels and experiments, limiting its widespread application.
Here we provide a proof-of-principle demonstration that the phototoxic effects of an imaging experiment can be identified directly from a single phase-contrast image using deep convolutional neural networks (ConvNets).
This lays the groundwork for an automated tool for assessing cell health in a wide range of imaging experiments.
Interpretability of such a method is crucial for its adoption.
We take steps towards interpreting the classification mechanism of the trained ConvNet by visualizing salient features of images that contribute to accurate classification. 


\end{abstract}


\end{frontmatter}


\section{Introduction}
\label{sec:intro}

In fluorescence microscopy, cells are exposed to high power illumination light. Excitation intensities on a standard widefield fluorescence microscope are on the order of 400\si{\watt\per\square\centi\metre}, about 2500-fold higher than the flux of sunlight hitting Earth on the brightest day of the year. Illumination intensity on a point scanning confocal may be as much as three orders of magnitude higher. In most cases, mammalian tissue culture cells are derived from tissues that are never exposed to light in their native context. Thu, this high-intensity illumination light damages cells, and has been shown to induce a wide variety of biological changes \cite{Gorgidze1998,Sparrow2001,Wagner2010,Knoll2015}. Some cellular responses to phototoxicity, such as retraction, rounding, or death by apoptosis, are visually striking even to researchers with little experience in live cell imaging, but many effects of high-intensity illumination are less obvious. 
By the time the obvious phenotypes appear, cells may already be experiencing the effects of photodamage. For example, microtubule dynamics \cite{Waldchen2015} and contractile dynamics \cite{Knoll2015} have been shown to be altered at relatively small light doses, before major, visually obvious, morphological changes are visible.
 
Phototoxicity can be minimized with carefully chosen acquisition settings, but choosing such appropriate settings requires a method for determining whether or not cells are experiencing photodamage. Most researchers detect phototoxicity during live cell imaging experiments by inspecting transmitted light images and looking for visibly sick cells. 
For example, cell biologists with many years of experience 
can learn to identify subtle signs of phototoxicity, using cues that may be easily missed by less experienced researchers.  
%
Since this qualitative method relies on the expertise of the experimenter, it is subject to bias and can significantly decrease the reproducibility of such studies.
 
Existing quantitative methods to measure phototoxicity are generally difficult to implement, requiring either additional fluorescent probes \cite{Dixit2003} or imaging of the sample hours or days after the initial experiment \cite{Carlton2010,Wagner2010,Konig1999}. Some methods calibrate phototoxicity using standard samples, and use relative measurements of phototoxicity to select minimally damaging imaging conditions \cite{Tinevez2012}. Because the effects of phototoxicity vary widely between samples, these methods are limited in their utility.
Thus there is a need for a quantitative and unbiased method to assess cell health and phototoxicity that can be easily applied to a wide range of microscopy samples.  We propose to address this issue by applying deep convolutional neural networks (ConvNets), which have recently achieved human-level accuracy for other image classification tasks \cite{alexnet,vgg,googlenet,resnet}, to assess phototoxicity in phase contrast transmitted light images.
 
We chose to work with phase contrast images because they
can be readily integrated into the workflow of many if not most light microscopy experiments. These images result in insignificant additional light exposure for the sample, and do not require the introduction of new labels. 
Being “label-free”, phase contrast images are also non-specific, their content being derived solely from variations in the optical path length through the sample resulting from local changes in index of refraction. This makes these images information rich, and therefore it's plausible that they contain features indicative of cellular phototoxicity response.

\begin{figure}
	\begin{minipage}{0.3\textwidth}
    	\begin{minipage}{0.9\textwidth}
	   		\caption{\textbf{Pipeline.} We propose an automated method for identifying phototoxocity in a live-cell imaging experiment, from a single transmitted light (TL) image taken at the end of the experiment.  A ConvNet classifies the TL image as sick or healthy, enabling the researcher to discard unreliable data and adjust imaging conditions, if necessary.}
			\label{fig:pipeline}
  		\end{minipage}
  	\end{minipage}
    \begin{minipage}{0.7\textwidth}
    	\begin{flushright}
   		\includegraphics[width=1.0\textwidth]
                   {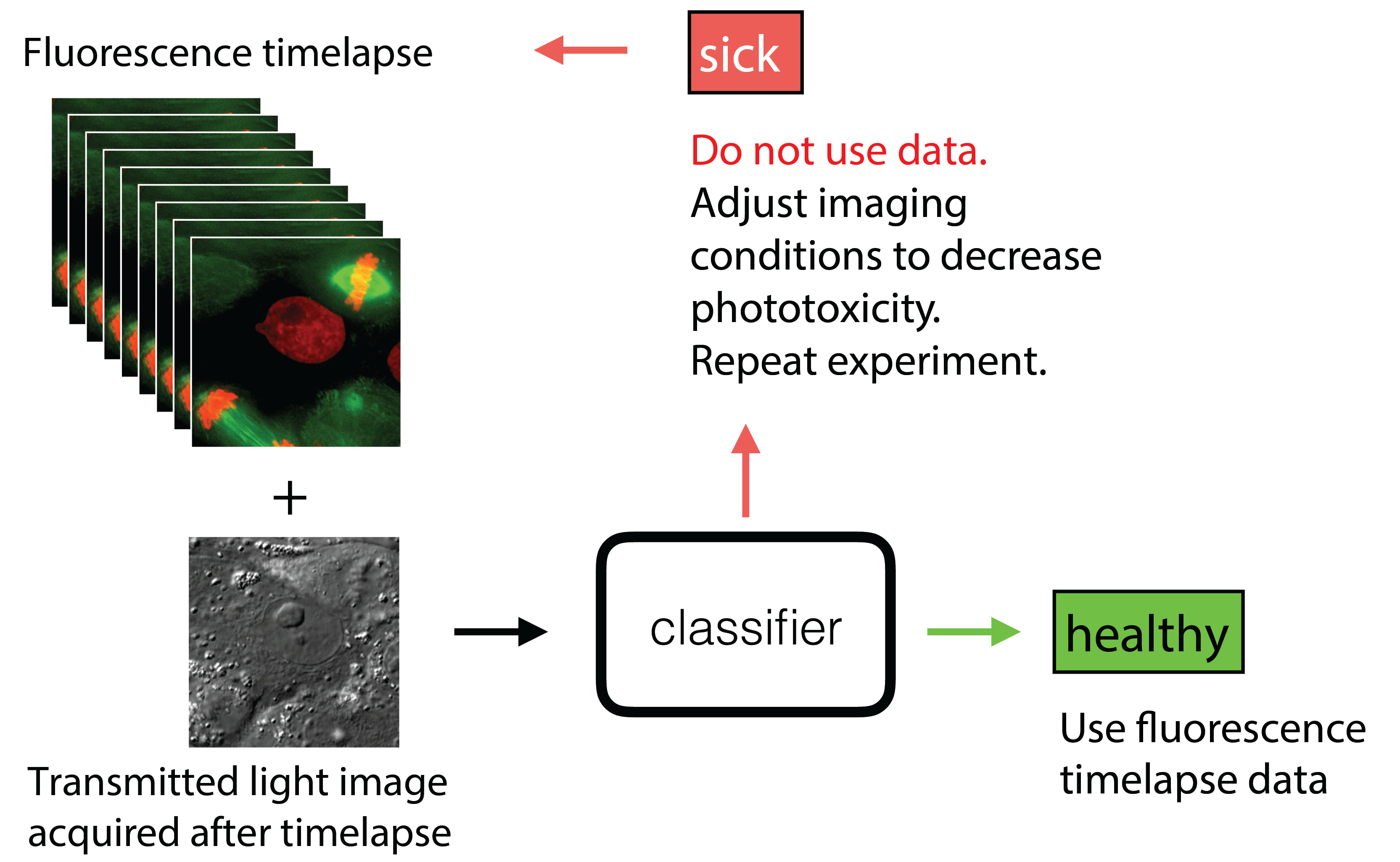}
    	\end{flushright}	
    \end{minipage}
    \label{fig:workflow}
\end{figure}


Furthermore, phase contrast images can be readily acquired in large quantities from cells that have experienced a known dose of fluorescence illumination light. These experimentally-derived annotations can then provide a supervisory signal for training a deep ConvNet. Once trained, this ConvNet provides a powerful, objective, and unbiased assessment of photoxicity, that can be easily applied to a single phase contrast image acquired at the end of an imaging experiment (see Figure~\ref{fig:workflow} for proposed workflow).

\section{Methods}
\label{sec:methods}

\subsection{Experimental Methods}
\label{subsec:expt}

U2OS cells were cultured in McCoy’s 5A media supplemented with 10\% FBS and penicillin-streptomycin, according to standard protocols. Cells were plated for imaging on black 24-well or 6-well \#1.5 coverslip glass bottom plates (CellVis).

All phototoxicity experiments were performed on a Nikon Ti Eclipse inverted microscope equipped with a Lumencor Spectra-X light engine for fluorescence illumination, a CoolLED pE-100 illuminator for transmitted light, Prior Proscan III motorized stage, and an Okolab cage incubator to maintain samples at 37$^{\circ}$C under humidified 5\% CO2. All experiments were performed with a 20x/0.75 Plan Apo Ph2 DM objective lens. Illumination to induce phototoxicity was delivered using the violet LED (395/30nm) in the Spectra-X light engine. Images were collected on a Hamamatsu ORCA-ER cooled CCD camera. The microscope was controlled with MetaMorph image acquisition software (Molecular Devices). 

For each experiment, 3-4 non-overlapping positions per well were chosen for imaging. 
All positions in a single well received the same light dose.
A single dose of continuous 395nm light was used to induce phototoxicity. 
A range of exposure times were used, from 0 to 200s.  
The 200s maximum was empirically determined to cause mass cell death by the end of the 18-24 hour timelapse.
Light exposure was automated with a custom MetaMorph journal, which moved the stage to each stored position and turned on the LED illuminator for 0-200s, depending on the specified light dose. A multi-position timelapse experiment was started immediately after light exposure using MetaMorph’s Multi-Dimensional Acquisition. Phase contrast images of each position were collected every 8 minutes for 18-24 hours. 

\subsection{Data Annotation}
\label{subsec:methods:annotation}

\textbf{Data Set A.} To generate the first data set for training and testing, we selected stage positions that could be labeled en masse.  Images of healthy cells were generated by simply selecting stage positions that were not exposed to any UV light.  To annotate sick cells, we first selected stage positions where every cell was unambiguously dead or dying by the end of the 18-24 hr timelapse.  From these positions, we searched for images at the beginning of the timelapse where cells were still relatively healthy looking, and these few frames were labeled as sick.  From each of these annotated full images (1344x1024 px), we extracted overlapping crops of 256x256 px, generating 54 annotated crops per full annotated image.  By this strategy, we were able to easily generate a large amount of annotated data (approx. 2000 crops for training); however, as a consequence, sick cells in this data set all correspond to relatively high UV exposure conditions.

\begin{figure*}
\begin{center}
   \includegraphics[width=1.0\textwidth]
                   {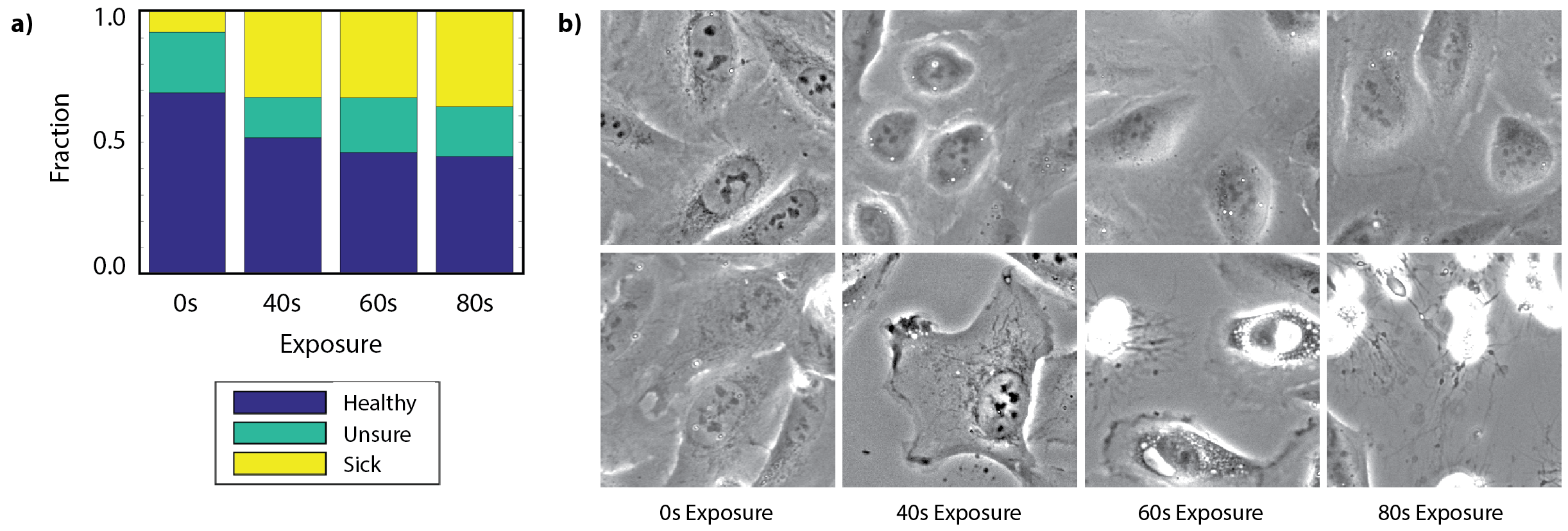}
\end{center}
\vspace{-10pt}
   \caption{\textbf{Expert Annotated Data.} Four cell biologists with extensive experience imaging live mammalian cells (``experts'') annotated images of cells, that were exposed to 395nm illumination light for varying periods of time, as ``Healthy'', ``Sick'' or ``Unsure'' if they were uncertain. The distribution of these annotations a) varied with light dose, along with the visual phenotype of cells from these conditions b), which was highly variable, containing cells with both relatively normal (top row) and unhealthy (bottom row) appearances.}
\label{fig:surveydata}
\end{figure*}

\textbf{Data Set B.} At numerous stage positions, generally corresponding to lower light dose, UV exposure caused local regions of cells to die, while other regions persisted, and even seemingly recovered from the initial insult.  
In such images, our previous mass-annotation strategy was not possible, yet many of these cells still displayed signs of phototoxicity identifiable by microscopists with extensive live-cell imaging experience. 
Thus, we employed human annotation by cell biologists (``experts'') who have used fluorescence and phase microscopy to image live mammalian cells in their own research for many years, and have acquired the knowledge necessary to identify phototoxicity based on commonly accepted indicators and their own experience. 
To generate the data for annotating, we selected images with 0-80s UV exposure and extracted 256x256 px non-overlapping crops.  Each crop contained 10 consecutive frames from the timelapse (temporal information is a strong discriminator between healthy and sick).  
To facilitate remote annotation, we constructed a javascript-based website that presented images at random.  
Experts were asked to categorize each 10-frame image sequence as ``Healthy'', ``Sick'' or ``Unsure'' and responses were logged in an SQL database (see Figure~\ref{fig:surveydata} for a summary of these annotations).  To ensure annotation overlap, every 5th image presented to the annotator was selected from a pool of images that were already scored by another annotator.  After an image region was annotated, all 10 frames were associated with the same class and used as ground truth data.  To avoid over-estimating the generalization capacity of our ConvNets, we were careful to ensure that the train and test sets did not contain crops from images acquired at the same stage position. After filtering these annotated images for concordance among our experts we had approx. 900 images, of which 700 were used for training.

\subsection{Pre-processing and Data Augmentation}
\label{subsec:preproc}

Data augmentation was used to expand the size of the training data set.  Each training image was augmented to generate multiple near copies, with the same class, via the following procedure:

\begin{enumerate}
\item Thin-plate spline warp (\textbf{w}): Input images were deformed by a thin-plate spline, as proposed by~\cite{unet}. Control points were placed on a grid at 128px intervals, and displaced according to $\mathcal{N}(0,\sigma^2_{disp} \mathbb{I})$, with $\sigma_{disp}=10$px.
\item Rotation and mirror image (\textbf{r}): All 8 unique combinations of horizontal and vertical reflections, with 90$^{\circ}$ rotations were applied.
\item Gaussian blur (\textbf{b}): Images were blurred using a Gaussian blur kernel with $\sigma_{blur}$ uniformly sampled from 1.1 - 1.5px to represent focus variation during imaging.
\item Crop (\textbf{c}): Images were cropped from their original size of 1344x1024 px down to 256x256 px, such that each cropped field of view contained approximately 1-6 uncentered cells. In the case of low light dosage, this allowed us to account for the fact that each full image may contain a combination of healthy and sick cells.  Overlapping crops were taken with a stride of 128px.
\item Time (\textbf{t}): A small window in time from the time-lapse was used, and all images were given the same label.
\item Additive Gaussian noise (\textbf{n}): 
After variance normalization, Gaussian noise was added with $\sigma_{noise}=0.5$.
\end{enumerate}

Note that in all cases, independent of augmentation by Gaussian noise, variance normalization (such that $\mu_{image}=0$, and $\sigma_{image}=1$) was applied at the level of cropped images.
Applying the above transformations, we generated a total of \textbf{w} $\cdot$ \textbf{r} $\cdot$ \textbf{b} $\cdot$ \textbf{c} = 3x8x2x54 = 2592 augmented images for every full sized image in Data Set A.  For Data Set B, our experts labeled cropped images, rather than full images, limiting the amount of augmentation that we could apply to each annotated image to \textbf{r} $\cdot$ \textbf{b} $\cdot$ \textbf{t} $\cdot$ \textbf{n} = 8x3x10x3 = 720 fold. The extent of augmentation was not optimized.

\subsection{Network Architecture and Hyper-parameters}
\label{subsec:methods:architecture}

The input to our network was a 220x220 px crop, randomly selected from within a 256x256 px image generated above. Note that this introduces additional augmentation. Our network architecture followed the VGG-style~\cite{vgg} with small (3x3 px), stacked convolutional kernels with stride and pad equal to 1 px. We used 8 convolutional (conv) layers, 3 fully connected (fc) layers, and 4 max pool layers.  We increased the number of feature maps by a factor of two after every spatial pooling layer, resulting in an overall bi-pyramidal architecture.  Each conv or fc layer was batch normalized~\cite{batchnorm}, with learned scale and shift, and then a ReLU non-linearity was applied \cite{nair2010rectified,glorot2011deep}.  The first two fc layers were regularized by drop-out, with a drop-out ratio of 0.5.  The final activation was a 2-way softmax classification.  For more details on this architecture, see Table~\ref{tab:architecture}.

We trained the networks using mini-batch stochastic gradient descent, based on backpropagation with Nesterov momentum. The loss function was cross-entropy, with additional regularization in the form of $L_2$ weight decay. All networks were trained from random initializations, except where explicitly stated, using the Xavier initialization scheme~\cite{xavier}.  The initial learning rate was set to 0.0003, and decayed according to an inverse schedule.  Momentum was 0.9, and weight decay was 0.001.
Training was done using a single GeForce TitanX (Maxwell) or GeForce GTX1080 (Pascal), for approx. 500k iterations (16hrs) with a batch size of 20 images.
For additional details and parameter values, the train and solver prototxts for the best performing networks can be found at \url{https://github.com/HMS-IDAC/DeadNet}.

\begin{table}
	\begin{center}
    	\begin{tabular}{|p{2.2cm}|p{2.5cm}|p{2.5cm}|p{2.8cm}|p{1.5cm}|} \hline
 	   	\textbf{layer} & \textbf{size in} & \textbf{size out} & \textbf{kernel, stride} \\ \hline
        
		Conv1\_1 & 220x220x1 & 212x212x16 & 9x9x1, 1 \\ \hline
 	   	Conv1\_2 & 212x212x16 & 212x212x16 & 3x3x16, 1 \\ \hline
 	   	MaxPool\_1 & 212x212x16 & 106x106x16 & 2x2x1, 2 \\ \hline
        
		Conv2\_1 & 106x106x16 & 106x106x32 & 3x3x16, 1 \\ \hline
 	   	Conv2\_2 & 106x106x32 & 106x106x32 & 3x3x32, 1 \\ \hline
 	   	MaxPool\_2 & 106x106x32 & 53x53x32 & 2x2x1, 2 \\ \hline

		Conv3\_1 & 53x53x32 & 53x53x64 & 3x3x32, 1 \\ \hline
 	   	Conv3\_2 & 53x53x64 & 53x53x64 & 3x3x64, 1 \\ \hline
 	   	MaxPool\_3 & 53x53x64 & 26x26x64 & 2x2x1, 2 \\ \hline

		Conv4\_1 & 26x26x64 & 26x26x128 & 3x3x64, 1 \\ \hline
 	   	Conv4\_2 & 26x26x128 & 26x26x128 & 3x3x128, 1 \\ \hline
 	   	MaxPool\_4 & 26x26x128 & 13x13x128 & 2x2x1, 2 \\ \hline

		fc\_1 & 13x13x128 & 1x1x512 & 13x13x128 \\ \hline
		fc\_2 & 1x1x512 & 1x1x512 & 1x1x512 \\ \hline
		score & 1x1x512 & 1x1x2 & 1x1x512 \\ \hline

		\end{tabular}
	\end{center}
  	\caption{\textbf{ConvNet Architecture. } This architecture was trained on both Data Set A and Data Set B.}
  	\vspace{-10pt}
	\label{tab:architecture}
\end{table}

\subsection{Network accuracy statistics}
\label{subsec:methods:accurracy}

To estimate uncertainties on model classification performance we used bootstrap sampling over 100 virtual batches \cite{virtbatch} each composed of a random set of 10 healthy and 10 sick cells. We then sampled from these classifications \SI{1e5} bootstrap samples with replacement and calculated $95\%$ confidence intervals using the bias-corrected accelerated percentile method \cite{efron1982jackknife}.

\subsection{Visualization Methods}
\label{subsec:methods:visualization}

\textbf{Grad-CAM.}
We used the method of Gradient Class Activation Mapping (Grad-CAM) to visualize the salient regions of images that contributed significantly to their classification~\cite{Selvaraju:2016ww}.  Grad-CAM is a generalization of Class Activation Mapping (CAM) to arbitrary ConvNet architectures~\cite{cam}.  A Grad-CAM image can be computed as follows.  For an input image, $I$,
the ConvNet generates $K$ feature maps $A_{u,v}^k$, for $(u,v) \in \Omega_{L}$, $k \in K$, of reduced spatial domain $\Omega_L$ at layer $L$ in the network.  The output layer of the network computes a score, $S_c$, for each class $c$, before applying softmax activation to produce a probability distribution over all classes.  The weight, or importance, of feature map $k$, contributing to the score of class $c$, for a particular image can be computed by Equation~\ref{eq:GradCAM1}.

\begin{equation}
\alpha_{c,k}= \frac{1}{\left | \Omega_L \right |} \sum_{u,v} \frac{\partial S_c}{\partial A^k_{u,v}}
\label{eq:GradCAM1}
\end{equation}

The Grad-CAM image, $I^c_{\mathbf{Grad-CAM}}$, then takes the weighted sum of the feature activation maps, and finally suppresses any negative values (Equation~\ref{eq:GradCAM2}).  We visualized Grad-CAM images from the final conv layer (\ie, L = Conv4\_2), leading to a coarse 26x26 px activation map, and up-sampled this to the full image resolution.  

\begin{equation}
I^c_{\mathbf{Grad-CAM}}=\left [ \sum_{k} \alpha_{c,k} A^k_{u,v}  \right ]_{+}
\label{eq:GradCAM2}
\end{equation}

We initially observed that the Grad-CAM images constructed from overlapping crops were not perfectly consistent in the overlapping regions.  This was in large part due to the weights in Equation~\ref{eq:GradCAM1} being computed on a per-crop basis.  To mitigate this effect, we computed average weights over the ensemble of all training images of class $c$, following Equation~\ref{eq:GradCAM3}.

\begin{equation}
\hat{\alpha}_{c,k}=\frac{1}{N_c} \sum_{i} \mathbbm{1}_c^{(i)} \alpha_{c,k}^{(i)}
\label{eq:GradCAM3}
\end{equation}

Where $i$ is an index over images in the training set, $\mathbbm{1}_c^{(i)}$ is a function indicating whether image $i$ is from class $c$, and $N_c=\sum_{i} \mathbbm{1}_c^{(i)}$ is the number of images of class $c$ in the training set.

\textbf{Class model visualization}
We also estimated class models~\cite{cam} to infer what, if any, larger-scale structures the model relied on for discrimination. We first tested the class model estimation using a VGG-16 model trained on ImageNet~\cite{vgg}, where we determined that, inspired by \cite{DeepDream}, an additional regularization term encouraging spatial continuity improved the subjective quality of large-scale structure in class models, giving the update rule:

\begin{equation}
I \leftarrow I + \epsilon \left ( \frac{\partial S_c }{\partial I} - \lambda_1 I  -\lambda_2 \nabla^2I   \right )
\label{eq:ClassModel1}
\end{equation}

Where $\lambda_1$ represents a penalty on the $L_2$ norm of the class model image $I$, and $\lambda_2$ is a smoothness term encouraging local pixel correlations. Initializing $I$ with zeros and applying this update rule using VGG-16, we subjectively determined the optimal step size and regularization parameters to be $\epsilon=10$, $\lambda_1=0.01$,$\lambda_2=0.001$ with 500 iterations, which resulted in interpretable class models for most ImageNet classes tested. We then used the same parameters to estimate class models from our networks, initializing $I$ to either zeros or to an image from the opposing class.

\section{Results and Discussion}
\label{sec:results}

\subsection{Performance}
\label{subsec:results:performance}

\begin{figure*}
\begin{center}
   \includegraphics[width=1.0\textwidth]
                   {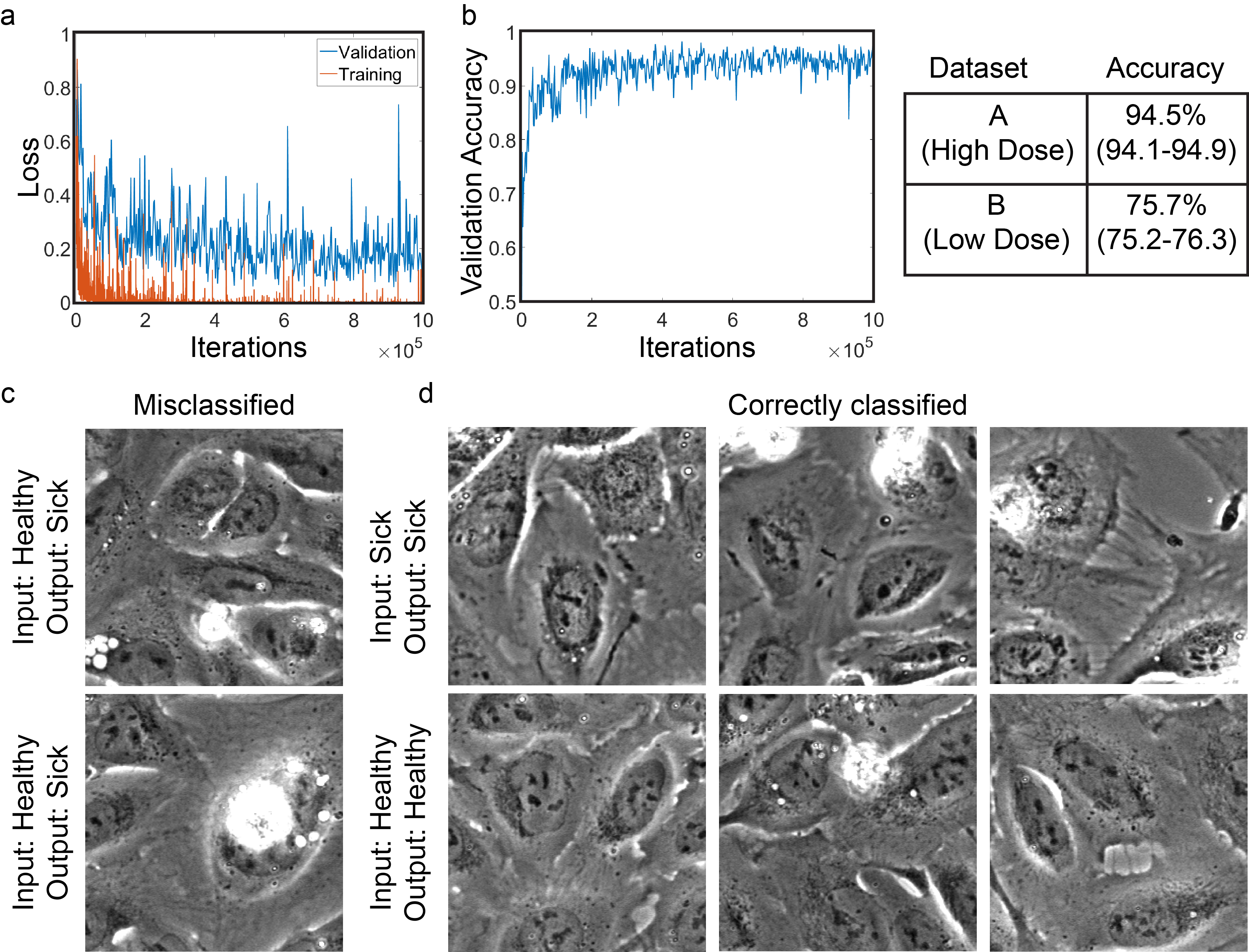}
\end{center}
   \caption{\textbf{Performance.} On dataset A the training and validation loss a) and validation accuracy b) improved rapidly but required several hundred thousand iterations to converge. The final accuracies (inset table) were much higher in Data Set A than in Data Set B. The failure modes c) in dataset A all consisted of healthy cells misclassified as sick, in some cases (lower panel) due to the presence of rounded cells in the healthy condition. Correctly classified cells d) had variable appearance in both classes, with often only subtle differences.}
\label{fig:results}
\end{figure*}

We trained ConvNets of identical architecture (see Table~\ref{tab:architecture}) on two different data sets, designated data set A and data set B, roughly representing low dose and high dose phototoxicity experiments, respectively.  On data set A, the ConvNet converged (both in terms of loss and accuracy on the test set) after approx. $500k$ iterations, and achieved a final accuracy of 94.5\% on the test set (Figure~\ref{fig:results}(a,b)).  This result corresponded to only seven incorrectly classified images out of 126, 4 of which were partially-overlapping crops.  We show two examples of healthy cells that were incorrectly labeled as sick by our ConvNet (Figure~\ref{fig:results}(c)), and in both cases it seems quite possible that the label is incorrect, not the prediction.  This could result from cell health being affected by factors other than phototoxicity.  We also show a number of examples of correctly classified cells (Figure~\ref{fig:results}(d)).  These images demonstrate the ability of the trained ConvNet to distinguish subtle changes between sick and healthy cells.

\begin{figure*}
\begin{center}
   \includegraphics[width=1.0\textwidth]
                   {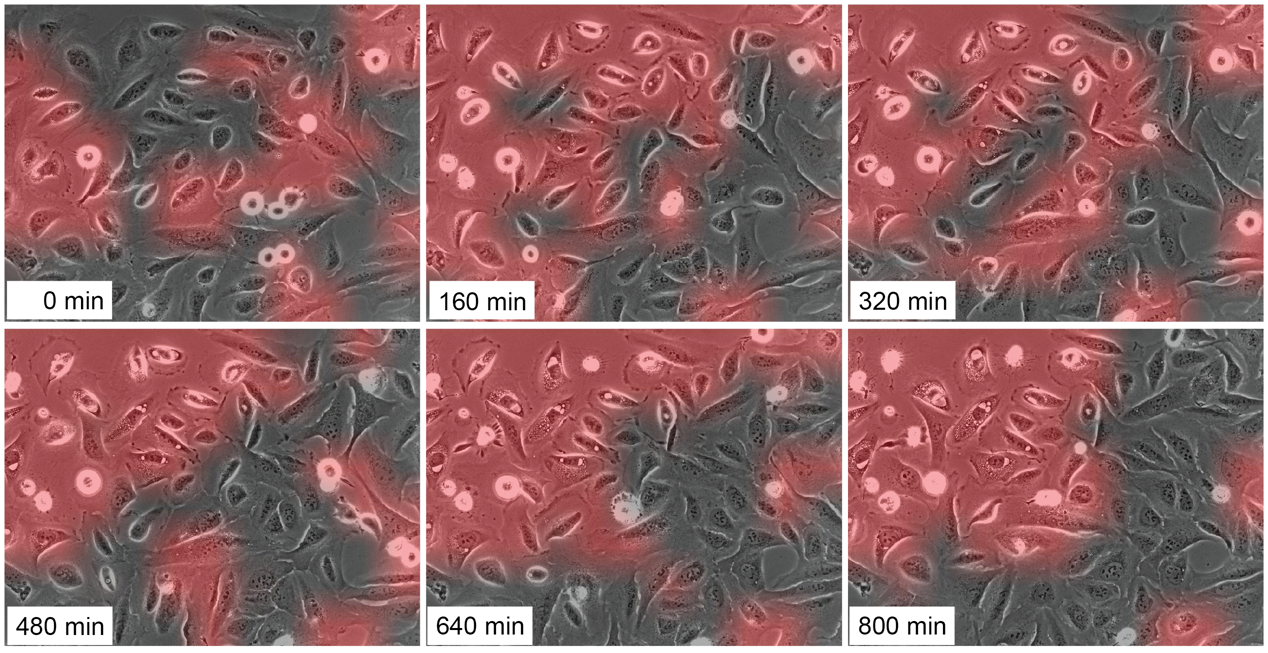}
\end{center}
\vspace{-10pt}
   \caption{\textbf{Heat maps.} We applied our classifier as a sliding window and then overlaid the resulting prediction on a timelapse image series, with bright red indicating a prediction of sick with high probability. The frames immediately after light exposure (upper left) show isolated unhealthy cells, with the fraction increasing over time (upper right panels) and with some cells eventually recovering and healthier cells entering the field of view (lower right panels).}
\label{fig:deathmaps}
\end{figure*}

After training this ConvNet, we applied it to a timelapse movie of cells that had been exposed to a lower dose of UV light.
The ConvNet was applied as a sliding window classifier, and the final classification heat map was then upsampled to the original image size via bilinear interpolation. The results clearly demonstrate the ability of the ConvNet to distinguish regions within the larger image where cells are sick and dying, from regions of healthy cells (Figure~\ref{fig:deathmaps}).  This highlights the applicability of our approach to analyzing large amounts of microscopy data, and automatically identifying unhealthy regions within this data.  In practice, images such as these with large swaths of cells showing clear signs of phototoxicity would be thrown out entirely, since all cells in the image were exposed to the same amount of light.

Next, we applied a similar strategy to train a ConvNet on data set B.  This data set in general corresponded to cells with lower light dose, requiring expert human annotators to generate the ground truth labels  (see Section~\ref{subsec:methods:annotation}). On this data set, we were able to achieve 75.7\% test set accuracy.  While this is significantly lower than the performance that we achieved on data set A, this data set contained numerous challenging and ambiguous samples. 

To determine a reasonable upper limit on the performance that we might expect from our ConvNet, we assessed the accuracy of the annotations by two methods. First, we presented a number of images to multiple annotators and measured their concordance on this subset. Out of 263 images with overlapping annotations, for which neither annotator labeled it as "unsure", there was disagreement on 49 (18.6\%) of these images. Before training our ConvNet, we filtered out all images with disagreeing labels; however, this doesn't imply that all remaining images were unequivocal.  Assuming that our annotators randomly agreed on equivocal images 50\% of the time (because there are two possible labels), this would imply that the original labeled set contained 37.2\% ambiguous images, and the final training set had 22.9\% ambiguous images remaining, after removing the disagreeing labels.  While this line of reasoning makes numerous assumptions, it does give some indication that the upper limit of performance that could be expected on the current training set is roughly 77.1\%, which is similar to the 75.2\% accuracy achieved by our ConvNet (Figure~\ref{fig:results}, inset table).

As a second indicator of expert human performance on this task,  we included a subset of images from cells that were never exposed to UV light.  We assume that these are true healthy cells, but acknowledge that they could show signs of poor health for other reasons (see Figure~\ref{fig:surveydata} for example images of unexposed cells). On this subset of images, our expert annotators labeled 230 / 256 (89.8\%) of these cells healthy.  On the subset of these same images that were not used during training, our ConvNet labeled 88.8\% of them as healthy (confidence interval = $88.1 - 89.5\%$)).  Thus, a ConvNet can be trained to achieve near expert level performance on this task.

\subsection{Visual Interpretation of ConvNet Behaviour}
\label{subsec:results:visualization}

As we have demonstrated, ConvNets can be trained to successfully distinguish U2OS cells suffering from phototoxicity.  In practice for this tool to be integrated into scientific imaging workflows, its interpretability is just as important as its accuracy.  
ConvNets are notoriously difficult to interpret, due to their high level of abstraction and end-to-end training.
A number of methods have recently been developed to address this concern, by producing visualizations that represent \textit{what} a trained net has learned to detect, and \textit{where} it is ``looking'' in an image when it makes a classification.

\begin{figure*}[!h]
\begin{center}
   \includegraphics[width=0.8\textwidth]
                   {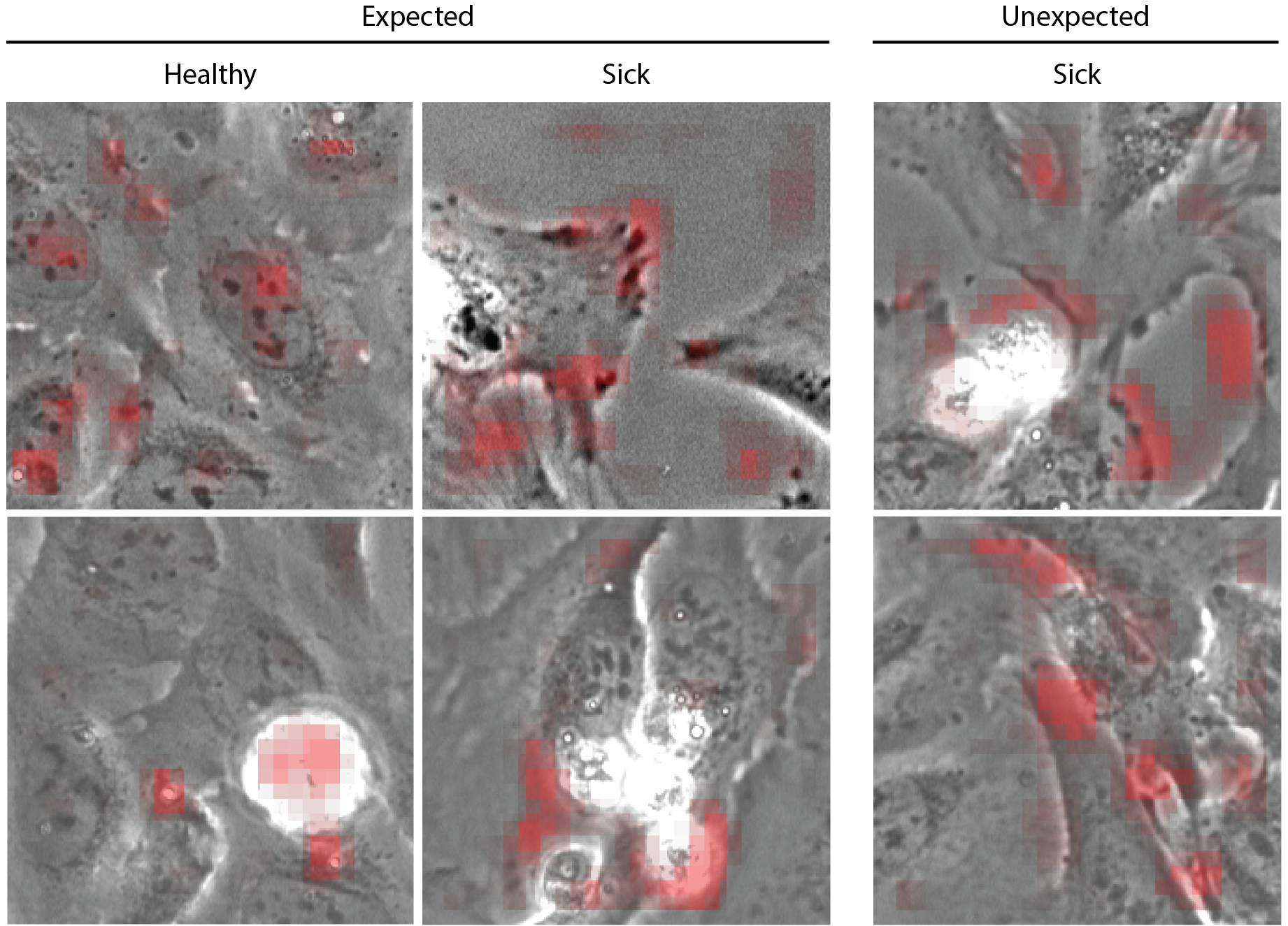}
\end{center}
\vspace{-10pt}
   \caption{\textbf{Class Activation Maps.} The Grad-CAM images of healthy cells often highlighted nucleoli (upper left panel) and large isolated vacuoles (lower left panel). In sick cells these highlighted retracting cell edges (top middle) and the edges of rounded up cells (lower middle). However in some cases these overlays revealed undesireable behavior such as the reliance on background areas (upper right) or simply highlighted structures that were not interpretably relevant to our experts (lower right).}
\label{fig:CAM}
\end{figure*}

We started by applying the method of gradient class activation mapping (Grad-CAM), to generate heat maps that highlight \textit{where} the ConvNet is looking~\cite{Selvaraju:2016ww}.  \Ie, these images highlight areas that are discriminative for a particular class (see Section~\ref{subsec:methods:visualization} for more details). The results of applying this method to the test data from data set A are shown in Figure~\ref{fig:CAM}.  Grad-CAM visualizations of data set B were more difficult to interpret, and aren't included here.

We observed that in healthy cells, 
the Grad-CAM images often highlighted regions with multiple nucleoli, large isolated vacuoles, and/or mitotic cells (Figure~\ref{fig:CAM} (left column)).  By contrast, structures that were highlighted in images of sick cells were dark aggregations at cell boundaries, retraction fibers, and/or the edge of rounded-up cells (Figure~\ref{fig:CAM} (middle column)).  In general, we observed more variability in the Grad-CAM images of sick cells than we did of healthy cells, and in many cases these images were uninterpretable by our expert microscopists (Figure~\ref{fig:CAM} (right column, bottom)).  This may imply that the ConvNet has learned to characterize healthy cells, and identified sick cells, which were more highly variable in phenotype, as outliers.  However, it is also possible that our visualizations were affected by over-fitting to the limited amount of data that these models were trained on.

Finally, we also occasionally observed interpretable, yet undesirable behavior in our ConvNets.  For example, empty regions of images often contributed to the classification of sick cells (Figure~\ref{fig:CAM} (right column, top)).  This is clearly interpretable, because as cells round up and die, it increases the exposed area in these images.  This highlights the importance of visualizing the behaviour of a ConvNet, for identifying and controlling unwanted confounding variation in the training data.

\begin{figure*}[!h]
\begin{center}
   \includegraphics[width=0.8\textwidth]
                   {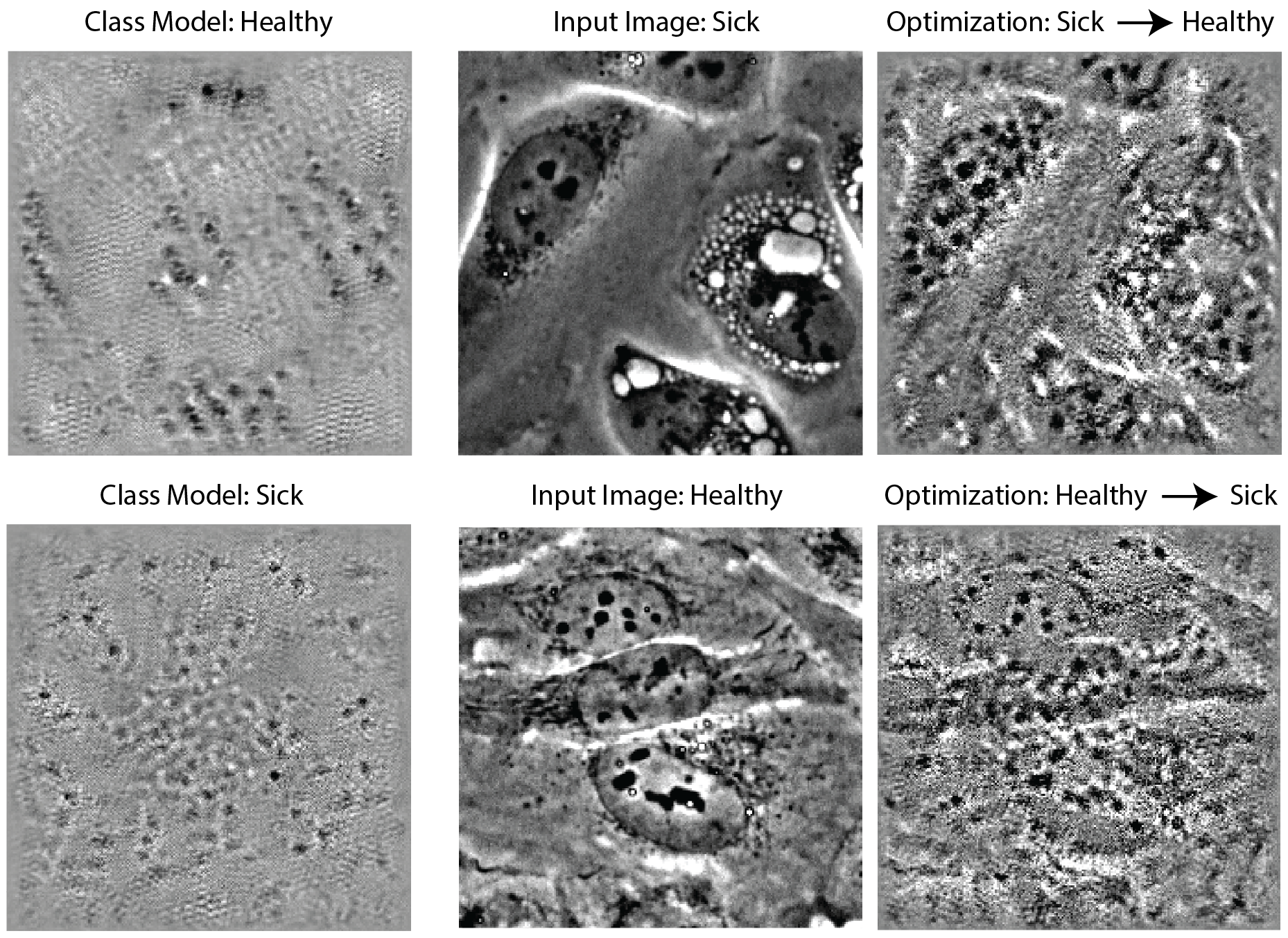}
\end{center}
\vspace{-10pt}
   \caption{\textbf{Class Model Visualization.} We estimated class models from either the healthy (top row) or sick (bottom row) classes. Initializing the model image $I$ to a zero image (left panels) produced structures weakly reminiscent of nucleoli for the healthy class (upper left panel) and more disorganized images for the sick class (lower left panel). Initializing $I$ to an image from the opposing class (right panels) added nucleoli to sick cells and removed vacuoles (upper right panel) but again produced more chaotic alterations when modeling the sick class (lower right panel). We interpret these changes cautiously, as some aspects of these alterations may be due to regularization.}
\label{fig:CMV}
\end{figure*}

To complement the Grad-CAM images, we also explored the use of class model visualizations~\cite{Simonyan:2013vv}, to visualize \textit{what} our trained ConvNets learned to look for.  This method generates images that optimize the score assigned to a particular class by the ConvNet, and in this way they reflect the learned model for the different classes (see Section~\ref{subsec:methods:visualization} for details).  The results of generating class model visualizations for healthy and sick classes, starting both from random initializations and from images of the opposing class, are shown in Figure~\ref{fig:CMV}.

Starting from a zero image, we observed that the ConvNet generated bright and dark punctae for both healthy and sick classes; however, with different spatial distributions.  Initializing this process from an image of the sick class (Figure~\ref{fig:CMV}, top right), and driving the optimization towards the healthy class, the ConvNet appeared to add nucleoli-like punctae in clusters near to the nucleus.  Starting from an image of the healthy class (Figure~\ref{fig:CMV} bottom right), and driving towards the sick class, the ConvNet added dark punctae more indiscriminately throughout the image.  We also observed that bright streaks at the edges of cells were reduced in both healthy and sick images, and multiple bright vacuoles were removed from the sick image; however removal of bright or dark regions was more difficult to interpret because it could be driven by the $L_2$ regularization term in the optimization process.

Overall, we found it difficult to interpret the class model visualizations generated by our trained ConvNets.  It's worth noting that this method does tend to generate very noisy images, even when applied to networks trained on millions of images~\cite{Simonyan:2013vv}.  Furthermore, we are applying this to a fine-grained recognition problem, where the classes are not as different as, say dog vs. car.  This may limit the applicability of such a visualization method.

We also observed that some of our trained ConvNets generated cleaner and more interpretable class model visualizations, despite their performance being otherwise indistinguishable to less interpretable ConvNets.  \Eg, the images shown in Figure~\ref{fig:CMV} were from a slightly different model architecture than described in Table~\ref{tab:architecture}, and trained on data set B.  This raises the interesting question of how to design and train ConvNets to optimize for their interpretability.  This question has similarly arisen in other works~\cite{cam}, but is potentially even more important and challenging to address in the scientific domain where interpretability is of utmost importance, and optimizing for it requires including experts in-the-loop.

\section{Conclusions}
\label{sec:conclusions}

In this paper, we have demonstrated the plausibility of using deep ConvNets to predict phototoxicity from label-free phase contrast images of U2OS cells.  On one data set, corresponding to a higher phototoxicity condition, we achieved 94.5\% prediction accuracy.  On a second, more challenging data set with lower phototoxicity, we obtained 75.7\% accuracy; however, this still rivaled expert level performance.  Going beyond this result will require larger annotated data sets, and we plan to pursue the use of live-dead markers to scale up our labeling efforts and more definitively assign class labels.  In future work, we also hope to generalize these findings to a wider range of cell types and microscopes, \textit{e.g.}, brightfield and DIC microscopy.

Another focus of this work was interpreting the predictions made by our ConvNets.  We applied methods for visualizing the internal class model that was learned by our ConvNets, and for highlighting features within an image that influenced its classification.  The ConvNet that achieved 94.5\% accuracy had fairly reproducible behaviour, highlighting nucleoli, large vacuoles and mitotic cells in healthy images, and retraction fibers, dark cell boundaries, and rounded up cells in sick images. By contrast, the class model visualizations were very challenging to interpret, likely due to the limited data that our model was trained on.  Interestingly, we found that interpretability varied almost orthogonally to performance, and we plan to further investigate network architectures that result in more interpretable behavior. Analysis of such networks would not only avoid classification based on undesireable confounding factors, but potentially provide novel insight into any generic cellular response to toxic levels of light.
Furthermore, it seems quite plausible that this type of analysis could generalize to other cellular insults.

\section{Acknowledgments}
\label{sec:acknowledgments}

The authors would like to acknowledge Yuyu Song, Ana Pasapera and Bob Fischer for contributing annotations to Data Set B. We would also like to acknowledge the Harvard Medical School Tools and Technology fund for supporting IDAC and this research.





\clearpage
{\small
\bibliographystyle{unsrt}
\bibliography{deadnet.bib}
}







\end{document}